\documentclass{article}
\usepackage{ijcai26}
\usepackage{natbib}

\usepackage{times}
\usepackage{soul}
\usepackage{url}
\usepackage[hidelinks]{hyperref}
\usepackage[utf8]{inputenc}
\usepackage[small]{caption}
\usepackage{graphicx}
\usepackage{amsmath}
\usepackage{enumitem}
\usepackage{amsthm}
\usepackage{booktabs}
\usepackage{algorithm}
\usepackage{algorithmic}
\usepackage{booktabs}
\usepackage{tabularx}
\usepackage{multirow}
\usepackage{makecell} 
\usepackage[switch]{lineno}
\usepackage{amssymb}

\title{Simulated Ignorance Fails: A Systematic Study of LLM Behaviors on Forecasting Problems Before Model Knowledge Cutoff}

\author{
    Zehan Li$^1$ \and
    Yuxuan Wang$^2$ \and
    Ali El Lahib$^2$ \and
    Ying-Jieh Xia$^2$ \and
    Xinyu Pi$^2$ \\
    \affiliations
    $^1$University of Chicago \\
    $^2$University of California, San Diego \\
    \emails
    zehan@uchicago.edu,
    \{waw009, aellahib, yix050, xpi\}@ucsd.edu
}

\begin{document}

\maketitle

\begin{abstract}
Evaluating LLM forecasting capabilities is constrained by a fundamental tension: 
prospective evaluation offers methodological rigor but prohibitive latency, 
while retrospective forecasting (RF)---evaluating on already-resolved events---faces rapidly shrinking clean evaluation data as SOTA models possess increasingly recent knowledge cutoffs. 
\textbf{Simulated Ignorance (SI)}, prompting models to suppress pre-cutoff knowledge, has emerged as a potential solution. 
We provide the first systematic test of whether SI can approximate \textbf{True Ignorance (TI)}. 
Across 477 competition-level questions and 9 models, we find that SI fails systematically: 
(1) cutoff instructions leave a 52\% performance gap between SI and TI; 
(2) chain-of-thought reasoning fails to suppress prior knowledge, even when reasoning traces contain no explicit post-cutoff references;
(3) reasoning-optimized models exhibit \textit{worse} SI fidelity despite superior reasoning trace quality. 
These findings demonstrate that prompts cannot reliably ``rewind'' model knowledge. 
We conclude that RF on pre-cutoff events is methodologically flawed; we recommend against using SI-based retrospective setups to benchmark forecasting capabilities.
\begin{figure}[!t]
\centering
\includegraphics[width=1\columnwidth]{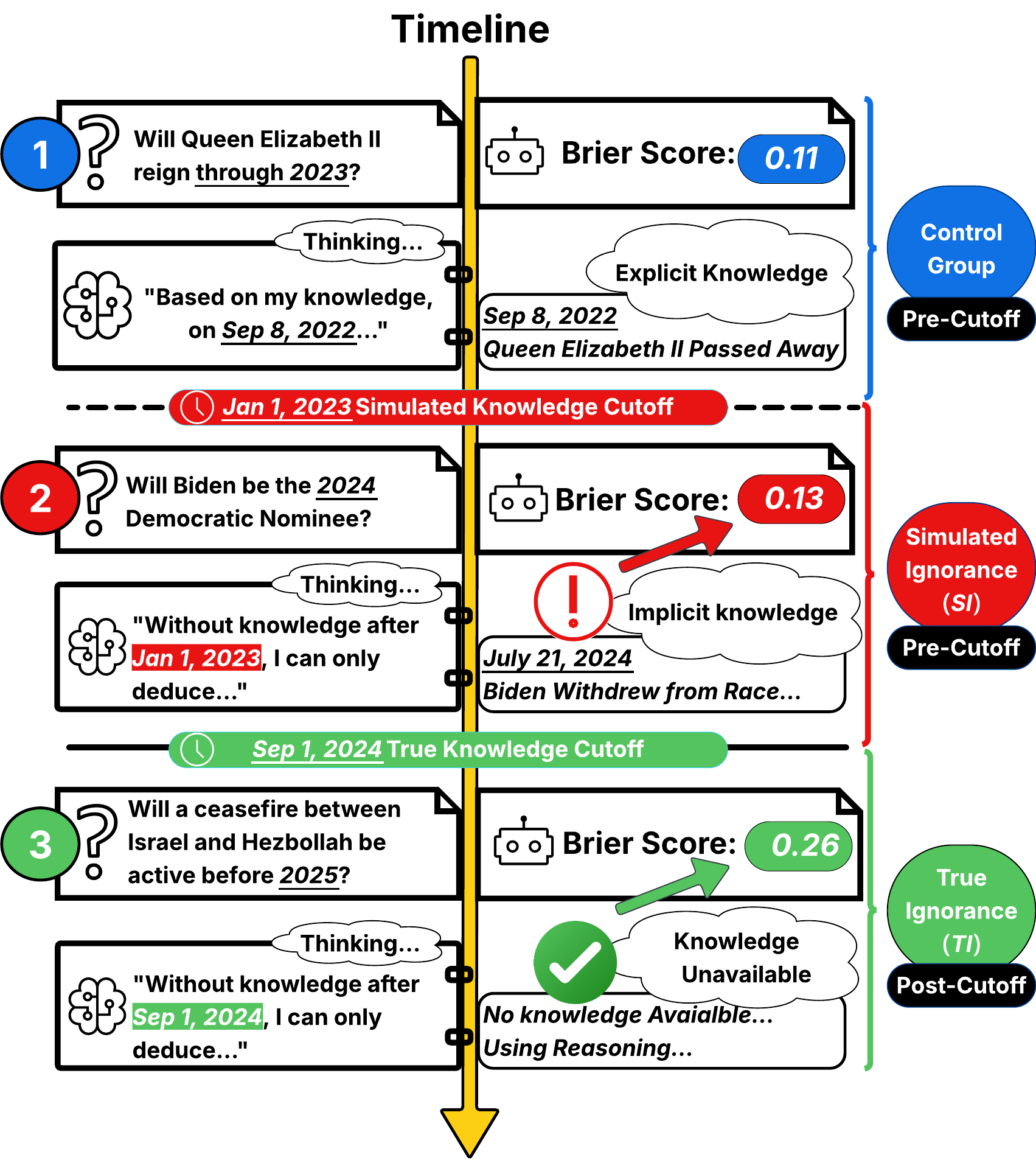}
\vspace{-2mm}
\caption{
        \textbf{Simulated Ignorance (SI) fails to suppress Implicit Knowledge.} 
        \textbf{(1) Control Group:} Models accurately predict past events using available \textbf{Explicit Knowledge} (Brier: 0.11). 
        \textbf{(2) Simulated Ignorance:} Despite a prompted cutoff (Jan 2023), the model retains \textbf{Implicit Knowledge} of future outcomes (e.g., Biden's withdrawal). As indicated by the red arrow, this leakage improves the forecast (Brier: 0.13)
        \textbf{(3) True Ignorance (TI):} When knowledge is genuinely \textbf{Unavailable} (e.g., post-training ceasefire), the model must rely on pure reasoning, resulting in significantly higher uncertainty (0.26). 
        The SI--TI gap confirms that prompts cannot reliably ``rewind'' the model's knowledge state.
    }
\label{fig:flowchart}
\end{figure}
\end{abstract}

\section{Introduction}

LLMs are increasingly deployed as forecasting tools, with applications spanning geopolitical prediction, financial analysis, and scientific discovery \citep{zou2022forecastingfutureworldevents}. 
The appeal is clear: models trained on vast corpora of human knowledge might internalize patterns that enable them to anticipate future developments. 
However, evaluating such forecasting capabilities presents a fundamental methodological challenge.

Prospective evaluation---deploying models to make predictions about genuinely future events and waiting for outcomes---offers the gold standard for assessing forecasting ability, but suffers from prohibitive latency. 
Years may pass before predictions can be validated, making rapid model evaluation impractical. 
Consequently, the field has largely converged on \textit{retrospective forecasting (RF)} as a compromise: evaluating models on historical events that have already resolved \citep{zou2022forecastingfutureworldevents,phan2024llms}. 
When such events postdate the model's training cutoff, the model operates under \textit{True Ignorance (TI)}---it genuinely lacks knowledge of the outcome \citep{schoenegger2024aiaugmentedpredictionsllmassistants, halawi2024approachinghumanlevelforecastinglanguage}.

However, TI-based evaluation is becoming increasingly difficult.
As \citet{paleka2025pitfalls} highlight, state-of-the-art models now possess knowledge cutoffs extending into late 2024 or early 2025.
This means most historical events have already been seen during training---leaving very few post-cutoff events available for evaluation.
The most capable models are precisely those with the most recent cutoffs, and therefore the fewest usable evaluation questions.
The problem is worse for model comparison: if we want to evaluate multiple models on the same questions, those questions must postdate \textit{all} models' cutoffs, further limiting the pool.

One proposed solution is to extend RF to pre-cutoff events by instructing models to suppress their knowledge---a strategy we term \textbf{\textit{Simulated Ignorance (SI)}}: rather than requiring events to genuinely postdate a model's training, we might instead prompt the model to ``forget'' what it knows---to simulate a state of ignorance about events it has actually encountered during training \citep{pawelczyk2024incontextunlearninglanguagemodels,gao-etal-2025-prompts}. 
For instance, to evaluate forecasting on the 2024 U.S. presidential election, one might instruct the model: ``\textit{use only knowledge available before January 1, 2023}.'' 
If successful, this approach would unlock vast repositories of historical data for forecasting evaluation, circumventing this data scarcity. 

But does \textbf{\textit{SI}} actually work? 
The test is straightforward: if prompting successfully suppresses prior knowledge, then forecasting performance under \textbf{\textit{SI}} should match performance under \textbf{\textit{TI}}. 
Conversely, if models perform significantly better on questions they ``should not know,'' they are likely benefiting from leaked knowledge, invalidating \textbf{\textit{SI}}-based RF.

Our paper provides the first systematic empirical test of this assumption.
Across $477$ competition forecasting questions evaluated on $9$ models---partitioned into pre-cutoff questions (evaluated under SI) and post-cutoff questions (evaluated under TI)---we demonstrate that \textbf{\textit{SI}} fails systematically.

Our investigation proceeds through three increasingly strong interventions. 
\textbf{(1)} We evaluate instruction prompting with explicit temporal constraints; models exhibit a \textbf{$52\%$ performance gap} between \textbf{\textit{SI}} and \textbf{\textit{TI}}, with substantially inflated accuracy and confidence on questions whose answers they were instructed to suppress. 
\textbf{(2)} We elicit structured reasoning, hypothesizing that explicit reasoning might force models to ``show their thinking'' and adhere to temporal constraints. 
While structured reasoning narrows the gap, it fails to close it.
More concerningly, models produce reasoning traces that appear fully compliant with the cutoff instruction, yet their forecasting performance still reflect leaked knowledge---suggesting that surface-level inspection of reasoning cannot reliably diagnose leakage.
\textbf{(3)} We test reasoning-optimized models trained with reinforcement learning, hypothesizing they might exhibit stronger instruction-following in temporal suppression tasks \citep{deepseekai2025deepseekr1incentivizingreasoningcapability}. 
Paradoxically, these models exhibit \textit{worse} SI fidelity despite superior reasoning quality and overall forecasting performance.

 All of our intervention attempts failed to rewind model knowledge. 
We conclude that \textbf{SI-based retrospective evaluation}---using prompts to simulate ignorance about pre-cutoff events---is \textbf{methodologically flawed}: 
\textbf{\textit{SI}} cannot approximate \textbf{\textit{TI}}, and we recommend against benchmarking to pre-cutoff events since knowledge leakage cannot be eradicated.

\section{Related Work}

\subsection{LLM Forecasting and Retrospective Evaluation}

A growing body of work evaluates LLM forecasting capabilities across domains including geopolitics, economics, and science \citep{zou2022forecastingfutureworldevents,pratt2024forecasting,schoenegger2024aiaugmentedpredictionsllmassistants}. 
These evaluations rely on retrospective setups---testing models on events that have already resolved---due to the impracticality of waiting for future results.

However, such setups vary widely in how they handle the risk of knowledge leakage.
Some studies restrict evaluation to post-cutoff events, ensuring models genuinely lack knowledge of results \citep{schoenegger2024aiaugmentedpredictionsllmassistants, halawi2024approachinghumanlevelforecastinglanguage}.
Others evaluate on pre-cutoff events, relying on temporal prompts to suppress model knowledge \citep{zou2022forecastingfutureworldevents}.
Recent benchmarks have quietly moved away from pre-cutoff evaluation, constructing prospective-only datasets without explicitly justifying this methodological shift \citep{karger2025forecastbench,paleka2025pitfalls,futuresearch2025benchfuturepastcastingbenchmark}.

This implicit retreat suggests growing unease with prompt-based knowledge suppression, yet the concern remains ungrounded: no prior work has directly tested whether SI can approximate TI, or quantified how badly it fails.
Most studies either assume SI works without validation, or silently avoid it.
Our work makes this implicit concern explicit, providing the first systematic empirical comparison between SI and TI.

\subsection{Temporal Contamination and Surface Compliance}
The failure of SI to approximate TI reflects a broader phenomenon: models' forecasts may benefit from leaked knowledge even when instructed to ignore it.
This connects to work on data leakage in NLP evaluation more generally \citep{sainz-etal-2023-nlp,balloccu-etal-2024-leak,magar-schwartz-2022-data}. 
In domains such as finance, prior studies demonstrate that models benefit from leaked knowledge under retrospective evaluation \citep{lopezlira2025memorization}. 
A related line of work investigates prompt-based ``unlearning'': whether instructions can suppress explicit mentions of known facts \citep{pawelczyk2024incontextunlearninglanguagemodels,gao-etal-2025-prompts}.

Our setting exposes a critical distinction: \textit{surface compliance} versus \textit{implicit knowledge retention}. 
A model may obediently avoid mentioning post-cutoff facts in its reasoning while its forecasting performance still reflects leaked knowledge. 
We therefore measure leakage via forecasting scores (Brier scores) rather than reasoning content.



\subsection{Reasoning Faithfulness and Post-Hoc Rationalization}

If surface compliance does not guarantee knowledge suppression, a natural question arises: can we trust reasoning traces to reveal when models are ``cheating''?
Chain-of-thought (CoT) prompting is widely adopted to improve both interpretability and performance \citep{wei2022chainofthought,kojima2022zeroshot}. 
However, a growing literature shows that reasoning traces can be misleading: models may produce convincing explanations that do not reflect their actual decision process \citep{turpin2023unfaithful,lanham2023measuringfaithfulness,paul2024making}. 

This concern is amplified for reasoning-optimized models (e.g., O1, R1), trained via reinforcement learning to produce extended traces \citep{openai2024o1systemcard,deepseekai2025deepseekr1incentivizingreasoningcapability}. \citep{chen2025reasoningmodelsdontsay} demonstrate unfaithful reasoning in such models on mathematical tasks, where errors in CoT can be detected by verifying logical consistency.
Forecasting offers no such verification: there is no ``correct'' reasoning path, so a cutoff-compliant trace may still mask leaked knowledge.

Our results provide forecasting-specific evidence for this concern.
Reasoning-optimized models exhibit the largest SI--TI gaps despite producing the cleanest traces---suggesting RL training teaches models to rationalize predictions rather than derive them transparently.

\section{Experimental Setup}
\label{sec:setup}

\subsection{Problem Formulation}

We test whether \textbf{\textit{Simulated Ignorance (SI)}}---prompting models to use only knowledge before a specified date---can make retrospective forecasting on pre-cutoff events behave like genuine forecasting on post-cutoff events \textbf{\textit{(True Ignorance, TI)}}. Each model $m$ has a reported knowledge cutoff $K(m)$. We prompt models with a simulated cutoff $C$ preceding all question resolutions, and test whether this eliminates outcome-level dependence: if SI works, models should perform equally on questions resolved before versus after $K(m)$.

\subsection{Data}

We evaluate 9 models on 470 resolved binary questions from Metaculus tournament archives (2023--2025) \citep{metaculus2025platform}. Models span two categories: 6 \textit{non-reasoning} models and 3 \textit{reasoning-optimized} models trained with reinforcement learning. Questions are stratified across six domains (Business, Economics/Finance, Social/institutional, Entertainment, Technology/innovation, Politics/Geopolitics) to ensure balanced coverage.

To ensure reliability, we run each model-question pair 3 times per condition and average the predictions. Sample sizes vary slightly across models (Table~\ref{tab:models}) due to occasional API failures or malformed outputs; we exclude such cases from analysis. We macro-average all metrics over questions.

\begin{table}[h]
\centering
\footnotesize
\begin{tabular}{lcc}
\toprule
\textbf{Model} & \textbf{Cutoff} & \textbf{$n$ (Pre/Post)} \\
\midrule
\multicolumn{3}{l}{\textit{Reasoning (R)}} \\
GPT-5.1 & Sept '24 & 201 / 276 \\
Gemini-2.5-Pro & Jan '25 & 201 / 276 \\
DeepSeek-R1 & Jan '25 & 201 / 276 \\
\midrule
\multicolumn{3}{l}{\textit{Non-Reasoning (NR)}} \\
Claude 3.5 Sonnet & Apr '24 & 201 / 274 \\
Gemini-2.5-Flash & Jan '25 & 201 / 273 \\
Llama 4 Maverick & Aug '24 & 201 / 274 \\
DeepSeek-V3.2-Exp & Dec '24 & 201 / 273 \\
OpenAI-OSS-120b & Jun '24 & 201 / 276 \\
OpenAI-OSS-20b & Jun '24 & 199 / 276 \\
\bottomrule
\end{tabular}
\caption{Model metadata. R = reasoning-optimized via RLVR; NR = standard instruction-tuned. Cutoffs from official documentation. Sample sizes vary slightly due to API failures.The Model knowledge cutoff date is available in \citep{wang2025llmcutoff}}
\label{tab:models}
\end{table}

\subsection{Question Partition}

To ensure all models are evaluated on the same questions, we use a \textit{fixed} partition rather than per-model splits.
Let $K_{\min}$ and $K_{\max}$ denote the earliest and latest knowledge cutoffs across all evaluated models.
Each question $q$ has resolution date $R(q)$, and we partition questions as follows:

\begin{itemize}[nosep,leftmargin=*]
\item \textbf{Pre-cutoff}: $R(q) < K_{\min}$. The outcome predates \textit{all} models' cutoffs; every model has likely seen the answer during training.
\item \textbf{Post-cutoff}: $R(q) > K_{\max}$. The outcome postdates \textit{all} models' cutoffs; no model has seen the answer.
\end{itemize}

When we add a cutoff instruction, Pre-cutoff questions become the \textbf{SI} condition (models must suppress knowledge they have), and Post-cutoff questions become the \textbf{TI} condition (models genuinely lack knowledge). The core test is whether performance under SI matches performance under TI.

This design ensures all models face the same question sets, but introduces a potential confound: models with later cutoffs are temporally closer to Post-cutoff question resolutions, and may benefit from more recent background context even without knowing answers.
To verify that observed SI--TI gaps reflect genuine knowledge leakage rather than temporal proximity effects, we conduct robustness checks examining performance as a function of distance between cutoff and resolution (\S\ref{sec:robustness}).

\subsection{Prompting Conditions}

We cross \textit{reasoning elicitation} with \textit{cutoff constraint} in a $2 \times 2$ design:
\begin{center}
\small
\begin{tabular}{lcc}
\toprule
& \textbf{No Cutoff} & \textbf{Cutoff at $C$} \\
\midrule
\textbf{Direct} & G1 & G1$'$ \\
\textbf{Reasoning (CoT)} & G2 & G2$'$ \\
\bottomrule
\end{tabular}
\end{center}
We fix $C = \texttt{2023-01-01}$, preceding all question resolutions. 

Under G1 and G2 (no cutoff instruction), we observe baseline performance on Pre-cutoff and Post-cutoff questions.
Under G1$'$ and G2$'$ (with cutoff instruction), Pre-cutoff questions become the SI condition and Post-cutoff questions become the TI condition.
This design allows us to isolate the effects of cutoff instructions (G1 vs.\ G1$'$) and reasoning elicitation (G1$'$ vs.\ G2$'$) on the SI--TI gap.

\subsection{Metrics}

Models output probability $p \in [0,1]$; ground truth is $y \in \{0,1\}$. We report Brier score $B(p,y) = (p-y)^2$ (lower is better; uninformed baseline $p=0.5$ yields $B=0.25$). 

Under cutoff instruction conditions (G1$'$, G2$'$), we define the \textbf{SI--TI Gap}:
\begin{equation}
\Delta_g = \mathbb{E}_{q \in \text{Post}}[B_g(q)] - \mathbb{E}_{q \in \text{Pre}}[B_g(q)]
\end{equation}
A large positive $\Delta_g$ indicates substantially better Pre-cutoff performance under SI---evidence that models fail to suppress leaked knowledge. If SI were effective, $\Delta_g \approx 0$.

\section{Results}
\label{sec:results}

We test whether prompt-based interventions can make SI approximate TI. Across 9 models and three intervention layers, we find that all interventions reduce explicit leakage but fail to close the SI--TI gap.

\subsection{Layer 1: Cutoff Instructions Reduce but Do Not Eliminate the Gap}
\label{sec:l1}

\begin{figure}[t]
\centering
\includegraphics[width=0.95\columnwidth]{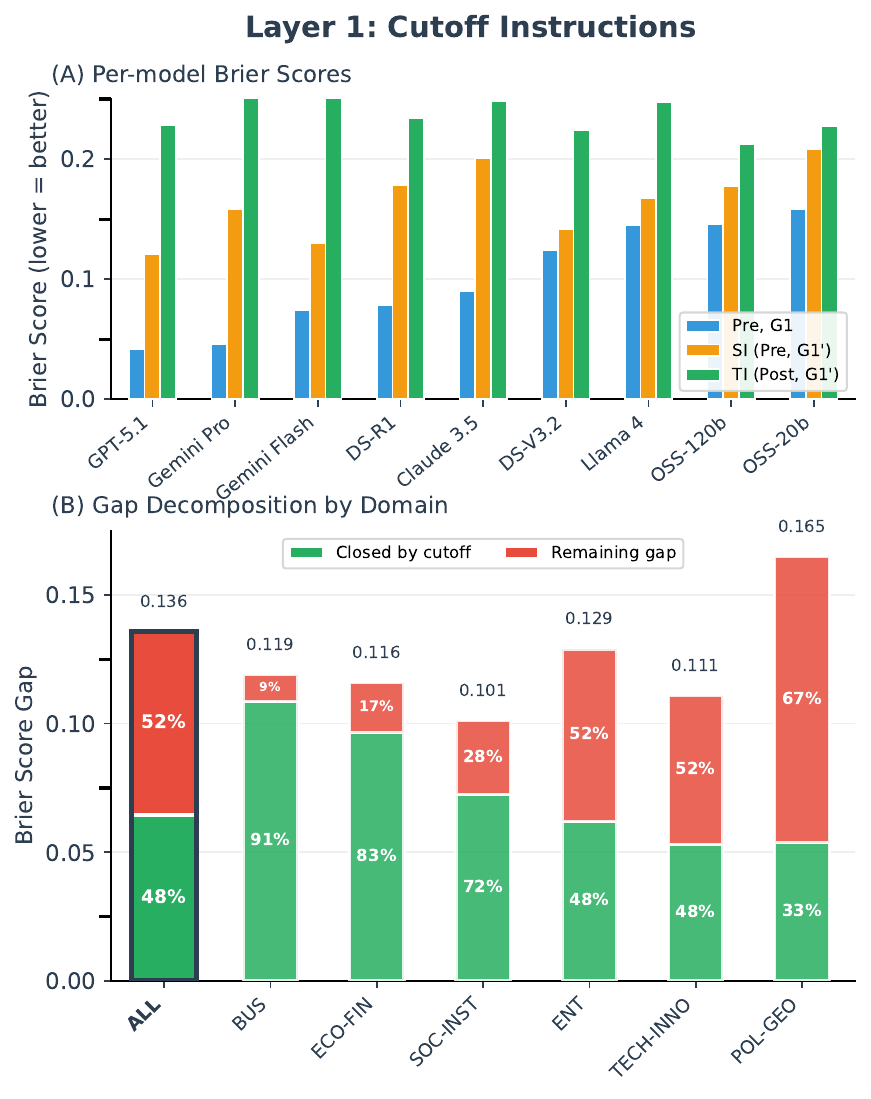}
\caption{\textbf{Layer 1: Cutoff instructions reduce but do not eliminate the SI--TI gap.} 
\textbf{(A)} Brier scores across 9 models under three conditions: Pre-cutoff baseline (Pre, G1), SI (Pre, G1$'$), and TI (Post, G1$'$). 
\textbf{(B)} Gap decomposition by domain: portion closed by cutoff (green) vs.\ residual SI--TI gap (red). 
Lower Brier = better.}
\label{fig:l1-cutoff}
\vspace{-3mm}
\end{figure}

Fig.~\ref{fig:l1-cutoff}A shows a consistent pattern across all 9 models: adding a cutoff instruction (G1$'$) reduces knowledge leakage on Pre-cutoff questions, moving it toward the Post-cutoff baseline---but a substantial gap remains. 
Quantitatively, the cutoff closes 48\% of the SI--TI gap, leaving 52\% unexplained (Fig.~\ref{fig:l1-cutoff}B).
All models still show significant gaps under the cutoff prompt.
SI therefore remains an unreliable proxy for TI even with explicit temporal constraints.

The SI--TI gap varies substantially across domains: cutoff instructions close 91\% of the gap in the most responsive domain (BUS) but only 33\% in the least responsive (POL-GEO).
This pattern suggests leakage severity correlates with event salience---geopolitical events like elections receive extensive coverage and become deeply encoded, making them harder to suppress than lower-salience business metrics (see supplementary material for detailed domain-level analysis).

\subsection{Layer 2: CoT Narrows the Gap but Clean Traces Hide Leakage}
\label{sec:l2}

We next test whether chain-of-thought (CoT) prompting helps models adhere to temporal constraints by forcing them to ``show their work.''

\begin{figure}[t]
\centering
\includegraphics[width=0.95\columnwidth]{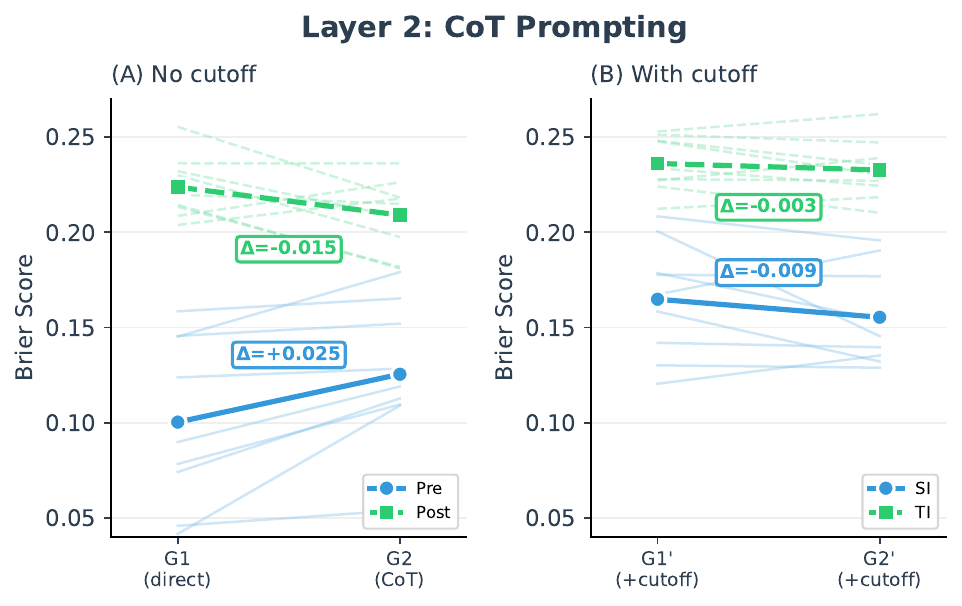}
\caption{\textbf{Layer 2: CoT prompting effects.}
(A) Without cutoff instruction: CoT worsens Pre-cutoff but improves Post-cutoff, narrowing the baseline gap.
(B) With cutoff instruction: G1$'$ $\to$ G2$'$ yields small changes on both SI and TI.}
\label{fig:l2-cot}
\end{figure}

\paragraph{CoT narrows the gap asymmetrically.}
Without a cutoff instruction (Fig.~\ref{fig:l2-cot}A), CoT improves Post-cutoff performance but \textit{worsens} Pre-cutoff performance, narrowing the baseline gap from both directions.
This asymmetry suggests that under direct prompting, models exploit a retrieval shortcut on Pre-cutoff questions---directly accessing memorized outcomes without explicit reasoning. 
CoT disrupts this shortcut by requiring verbalization, degrading Pre-cutoff performance while benefiting Post-cutoff where no shortcut exists.

With a cutoff instruction (Fig.~\ref{fig:l2-cot}B), switching from direct to CoT prompting (G1$'$ $\to$ G2$'$) produces only small changes on both SI and TI.
The cutoff instruction appears to already disrupt the retrieval shortcut, leaving less room for CoT to further reduce leakage.

\begin{figure}[t]
\centering
\includegraphics[width=0.95\columnwidth]{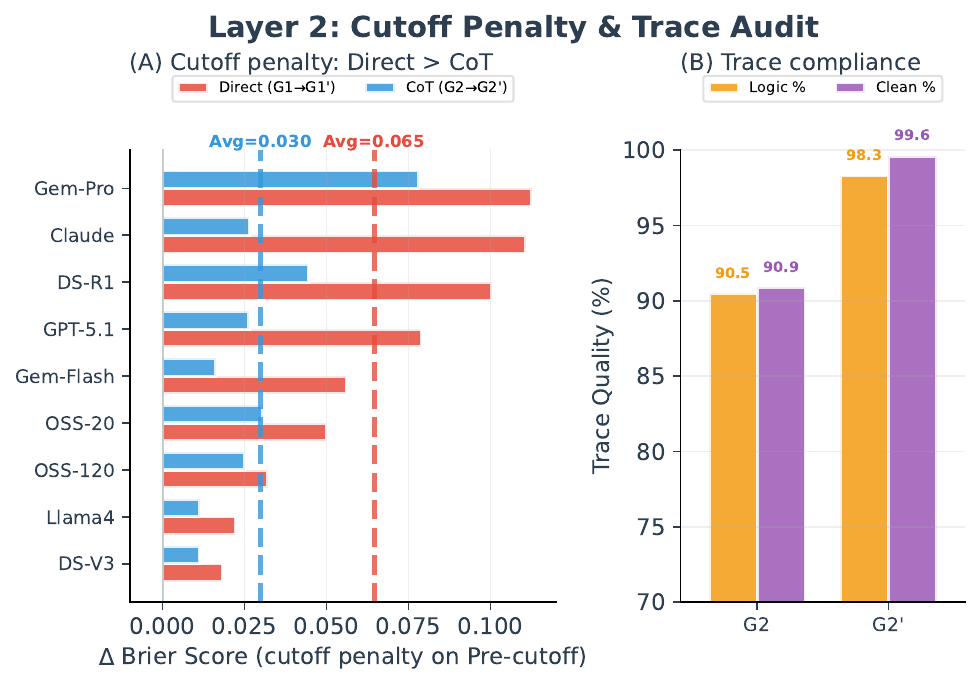}
\caption{\textbf{Layer 2: Trace compliance.}
(A) Cutoff penalty on Pre-cutoff is larger for direct prompting than CoT.
(B) Trace audit metrics improve from G2 to G2$'$.}
\label{fig:l2-trace}
\vspace{-3mm}
\end{figure}

\paragraph{Clean traces do not imply no leakage.}
We audit CoT traces on Pre-cutoff questions for logical consistency and explicit post-cutoff references.\footnote{Logic consistency checks for internal contradictions, mathematical errors, and reasoning-conclusion alignment. Cutoff compliance detects references to information only knowable after the specified cutoff. See supplemental material for more details.}Trace quality improves from G2 to G2$'$ (Fig.~\ref{fig:l2-trace}B): cutoff prompts make traces \textit{look} cleaner.
However, cleaner traces coexist with residual SI--TI gaps---the absence of explicit temporal violations does not guarantee the absence of leaked knowledge in the forecast.

\subsection{Layer 3: Reasoning-Optimized Models Show Cleaner Traces but Larger Gaps}
\label{sec:l3}

Reasoning-optimized models (e.g., O1, R1) are trained via reinforcement learning to produce structured reasoning and follow instructions consistently.
We test whether this training improves SI fidelity.

\begin{figure}[t]
\centering
\includegraphics[width=\columnwidth]{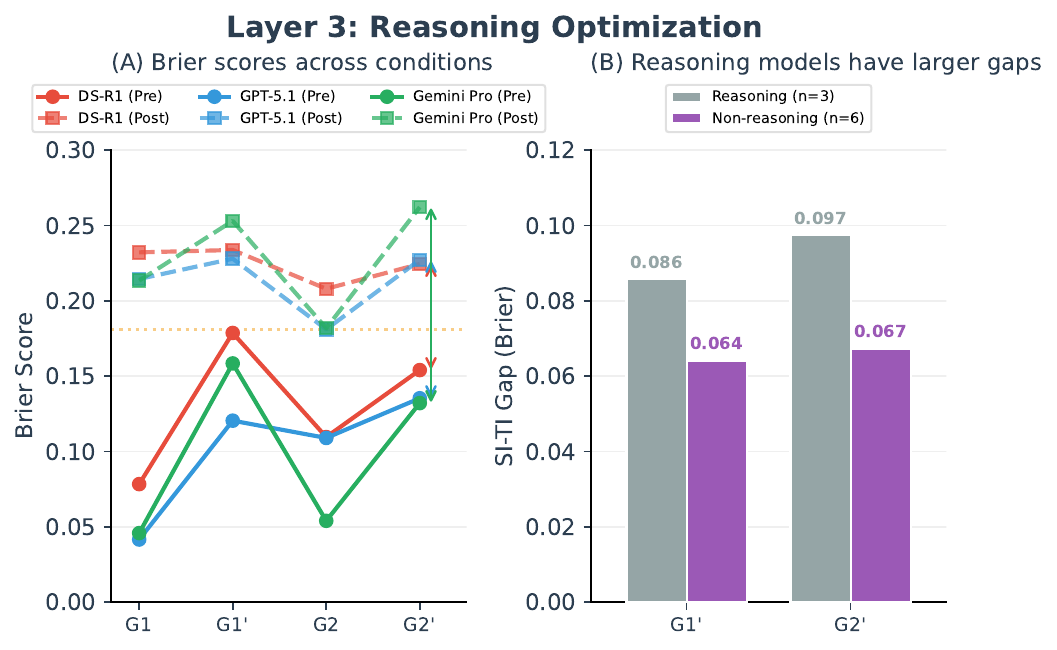}
\caption{\textbf{Layer 3: Reasoning-optimized models show cleaner traces but larger SI--TI gaps.}
(A) Brier scores across prompting conditions.
(B) SI--TI gap comparison: reasoning-optimized vs.\ non-reasoning models.}
\label{fig:l3-reasoning-models}
\end{figure}

\paragraph{Better TI performance, worse SI fidelity.}
Fig.~\ref{fig:l3-reasoning-models}A shows that reasoning-optimized models achieve strong TI performance across all prompting conditions.
Their traces also appear more compliant: under G2$'$, they show higher logic-consistency (99.4\% vs.\ 96.7\%) and cutoff-compliance (98.8\% vs.\ 98.2\%) than non-reasoning models (Table~\ref{tab:l3-trace-gap}).

However, reasoning-optimized models exhibit \textit{larger} SI--TI gaps than non-reasoning models (Fig.~\ref{fig:l3-reasoning-models}B): 0.086 vs.\ 0.064 under G1$'$, and 0.097 vs.\ 0.067 under G2$'$.
This dissociates forecasting capability from leakage resistance: models can forecast well under TI while failing to suppress prior knowledge under SI.

This pattern is consistent with RL training creating tension between two objectives: producing compliant traces and arriving at correct answers.
Under SI, models learn to construct reasoning that avoids explicit violations while still guiding toward the encoded answer---appearing compliant without being compliant (see supplementary for extended analysis).


\begin{table}[t]
\centering
\small
\begin{tabular}{lcc}
\toprule
Group & Logic (\%) & Clean (\%) \\
\midrule
Reasoning-optimized & 99.4 & 98.8 \\
Non-reasoning       & 96.7 & 98.2 \\
\bottomrule
\end{tabular}
\caption{\textbf{Layer 3: Trace audit under G2$'$.}
Reasoning-optimized models produce more coherent and cutoff-clean traces under G2$'$.}
\label{tab:l3-trace-gap}
\vspace{-3mm}

\end{table}
\subsection{Summary}
\label{sec:results-summary}

Across all three interventions, we observe a common failure mode: interventions operate at the level of \textit{model outputs}---what the model says---rather than \textit{model knowledge}---what the model knows. 

Cutoff instructions suppress explicit temporal references, CoT prompting produces coherent reasoning traces, and reasoning-optimized training yields high trace compliance---yet all three leave residual SI--TI gaps. A model can avoid every detectable form of leakage while implicit knowledge still influences its forecasting performances.

\section{Robustness \& Qualitative Analysis}
\label{sec:robustness}

Our main finding---that SI fails to approximate TI---could be challenged by an alternative explanation: perhaps post-cutoff questions are simply \textit{harder} to forecast, independent from data leakage. We address this concern with experiments.

\subsection{Difficulty Control: The Gap Is Model-Specific, Not Calendar-Specific}
\label{sec:difficulty}

If difficulty (not leakage) causes the SI--TI gap, all models should struggle on the same questions at the same calendar time.
If leakage causes the gap, performance should degrade at each model's \textit{own} knowledge cutoff, regardless of calendar time.
We test this with three complementary analyses.

\paragraph{Cross-model natural experiment.}
We exploit the fact that different models have different cutoffs.
Q4 2024 questions provide a decisive test: they are \textbf{Pre-cutoff} for DeepSeek-V3.2 (cutoff: Jan 2025) but \textbf{Post-cutoff} for GPT-5.1 (cutoff: Sept 2024).
If these questions were inherently harder, \textit{both} models should struggle.

\begin{table}[h]
\centering
\small
\begin{tabular}{llcc}
\toprule
\textbf{Model} & \textbf{Q4'24 Status} & \textbf{Brier} & \textbf{Horizon} \\
\midrule
GPT-5.1 (Sept'24) & TI & .204 & 1--3 mo \\
DeepSeek-V3.2 (Jan'25) & SI & .136 & 1--3 mo \\
\bottomrule
\end{tabular}
\caption{\textbf{Cross-model natural experiment.} Same questions, same horizon, different SI/TI status. The gap (+0.068) tracks knowledge cutoff, not question difficulty.}
\label{tab:cross-model}
\end{table}

The same questions yield Brier 0.136 for DeepSeek but 0.204 for GPT-5.1.
The only variable is whether the questions fall before or after each model's cutoff---strong evidence that the gap reflects leakage, not difficulty.
Importantly, both models achieve similar Brier scores on TI questions (GPT-5.1: 0.21; DeepSeek: 0.23), confirming that the difference on Q4 2024 questions is not due to overall model capability.

\paragraph{Temporal proximity does not explain the gap.}
One might worry that models with later cutoffs perform better on TI simply because they have more recent background context.
However, the cross-model experiment controls for this: both models face Q4 2024 questions at the same temporal distance from resolution, yet only the model for which these questions are SI shows elevated performance.

\begin{figure}[t]
\centering
\includegraphics[width=\columnwidth]{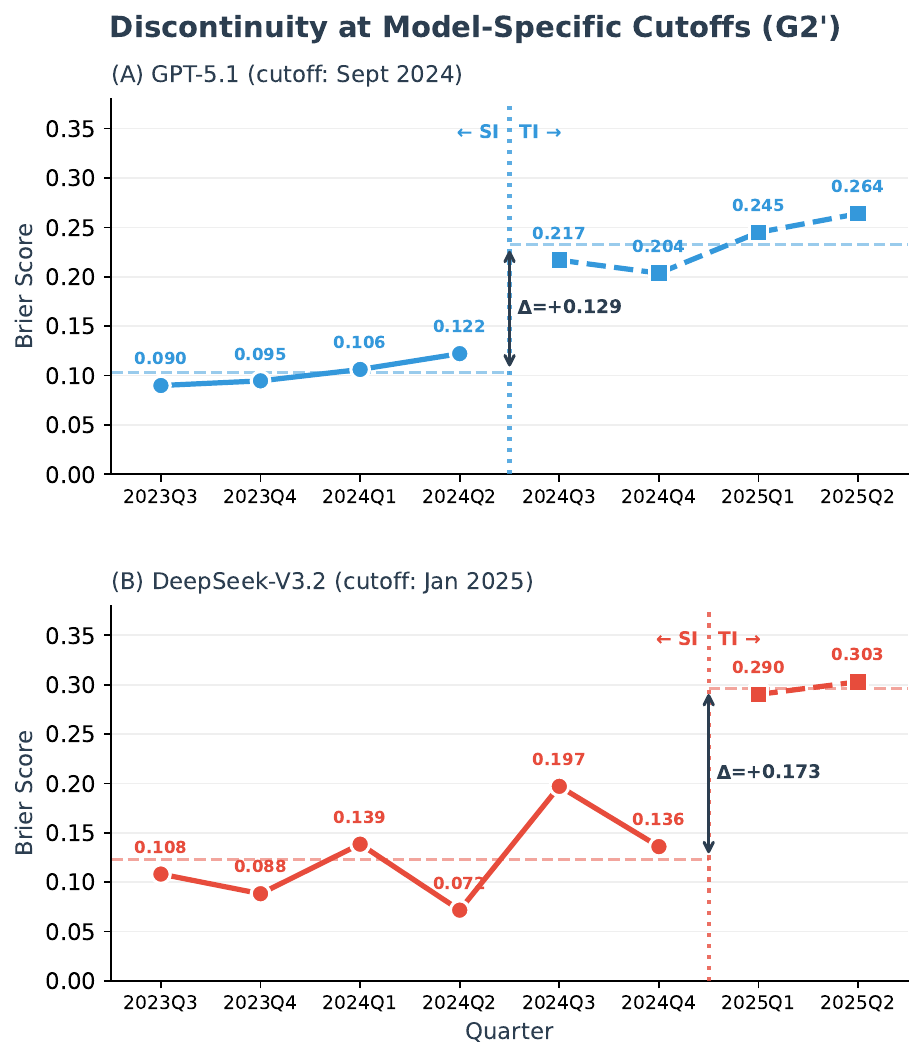}
\caption{\textbf{Discontinuity at model-specific cutoffs (G2$'$).} Each model shows a sharp Brier increase at its own cutoff (vertical line), not at a common calendar point. (A) GPT-5.1 cutoff: September 2024. (B) DeepSeek-V3.2 cutoff: January 2025.}
\label{fig:discontinuity}
\vspace{-1mm}
\end{figure}

\paragraph{Discontinuity at model-specific cutoffs.}
Figure~\ref{fig:discontinuity} shows quarterly Brier scores for two models.
If difficulty drove the gap, we would expect smooth degradation over calendar time.
Instead, each model shows a sharp jump at \textit{its own} cutoff:
GPT-5.1 jumps at Sept 2024 ($\Delta \approx +0.13$); DeepSeek jumps at Jan 2025 ($\Delta \approx +0.17$).
The discontinuity occurs at different calendar points for different models, ruling out calendar-time difficulty.

\paragraph{Human baseline confirms equal difficulty.}
To independently verify that SI and TI questions have comparable difficulty, we benchmark Metaculus community forecasters on both splits.
To avoid hindsight bias, we use only predictions from the first two weeks of each question.

\begin{table}[h]
\centering
\small
\begin{tabular}{lcc}
\toprule
\textbf{Question Split} & \textbf{Human Brier} & \textbf{Model Brier (avg)} \\
\midrule
Pre-cutoff (SI for models) & 0.232 & 0.11 \\
Post-cutoff (TI for models) & 0.243 & 0.24 \\
\midrule
$\Delta$ & \textbf{0.011} & \textbf{0.13} \\
\bottomrule
\end{tabular}
\caption{\textbf{Human baseline.} Humans show negligible gap between 
Pre- and Post-cutoff questions ($\Delta = 0.01$); models under SI/TI 
conditions show large gaps ($\Delta \approx 0.13$).}
\label{tab:human-baseline}
\end{table}

Humans find both splits equally difficult ($\Delta = 0.011$), while models show gaps an order of magnitude larger ($\Delta \approx 0.13$).
This confirms that the model gap reflects leaked knowledge, not question difficulty.

\subsection{Trace Audit Validation}
\label{sec:trace-audit}

We validate our automated (GPT-4o) trace audit against independent human 
annotations on a stratified sample ($n=36$), evaluating both logic 
consistency and cutoff compliance.

\begin{table}[h]
\centering
\small
\begin{tabular}{lcc}
\toprule
\textbf{Dimension} & \textbf{$\kappa$} & \textbf{Agreement} \\
\midrule
Logic consistency & 0.64 & 94\% \\
Cutoff compliance & 0.86 & 94\% \\
\bottomrule
\end{tabular}
\caption{\textbf{Audit validation.} High human-GPT agreement confirms 
our audit reliably detects explicit violations.}
\label{tab:human-agreement}
\end{table}

The strong agreement ($\kappa = 0.86$ for cutoff compliance) confirms 
that the persistent SI--TI gap under high trace compliance reflects 
\textit{implicit} knowledge retention, not failures in our audit methodology.

\subsection{Qualitative Analysis: The Illusion of Reasoning}
\label{sec:qualitative}

Our trace audit detects explicit violations but cannot capture 
\textit{implicit} knowledge that influences predictions without 
overt leakage. We present paired case studies (Table~\ref{tab:qualitative_cases}) 
under G2$'$ (CoT with cutoff instruction) where structurally similar 
questions yield dramatically different accuracy.

\begin{table*}[t]
\centering
\small
\resizebox{\textwidth}{!}{%
\begin{tabular}{l p{5.5cm} l c c p{6.5cm}}
\toprule
\textbf{Period} & \textbf{Question (Condensed)} & \textbf{Model} & \textbf{Pred ($p$)} & \textbf{Truth ($y$)} & \textbf{Reasoning Trace (CoT) Snippet \& Analysis} \\
\midrule
\multicolumn{6}{l}{\textit{\textbf{Case 1: Ceasefire Predictions} (Both are binary ceasefire-by-date questions with similar geopolitical framing)}} \\
\cmidrule(lr){1-6}
\multirow{2}{*}{\textbf{Pre (SI)}} & Q13959: Will Russia and Ukraine have a lasting ceasefire by Jan 1, 2024? & Gemini-Pro & 0.04 & 0 & \textit{``Looking at historical conflicts... lasting cease-fire within a year is relatively rare. A rough anchor... might be around 10--20\%.''} \\
& & & & & \textbf{$\rightarrow$ Accurate via historical priors.} \\
\addlinespace[0.1cm]
\multirow{2}{*}{\textbf{Post (TI)}} & Q30269: Will Israel and Lebanon have a ceasefire by Jan 1, 2025? & Gemini-Pro & 0.45 & 1 & \textit{``The Nov 25, 2024 report indicated a 60-day deal was close... $P(\text{deal}) \times P(\text{holds}) = 0.70 \times 0.60 = 0.42$.''} \\
& & & & & \textbf{$\rightarrow$ Similar reasoning framework, but results in hallucinated precision.} \\
\midrule
\multicolumn{6}{l}{\textit{\textbf{Case 2: International Agreements} (Identical task: predicting treaty signing)}} \\
\cmidrule(lr){1-6}
\multirow{2}{*}{\textbf{Pre (SI)}} & Q13986: In 2023 will any new country join NATO? & Gemini-Flash & 1.00 & 1 & \textit{``Turkey's parliament has ratified... Hungary's parliament also ratified... All other members have already ratified.''} \\
& & & & & \textbf{$\rightarrow$ Perfect score driven by precise factual recall.} \\
\addlinespace[0.1cm]
\multirow{2}{*}{\textbf{Post (TI)}} & Q37397: Will a country sign the Artemis Accords in June 2025? & Gemini-Flash & 0.67 & 0 & \textit{``Steady increase in signatories... $\sim$1 signing per month. Using Poisson distribution... $1 - e^{-1} = 0.633$.''} \\
& & & & & \textbf{$\rightarrow$ Rigorous logic (Poisson), but fails without leaked outcome.} \\
\bottomrule
\end{tabular}%
}
\caption{\textbf{Qualitative Case Studies.} Paired comparison of Pre-cutoff (SI) vs.\ Post-cutoff (TI) questions. In both cases, models use structurally similar reasoning (historical anchors or statistical models). However, Pre-cutoff predictions benefit from implicit knowledge (reflected in high confidence/accuracy), while Post-cutoff predictions using the same logic fail to capture ground truth.}
\label{tab:qualitative_cases}
\vspace{-2mm}
\end{table*}

\paragraph{Case Study 1: Ceasefire Predictions.}
Both questions require predicting a ceasefire approximately one month 
before the deadline. For the Pre-cutoff event (Russia-Ukraine), models achieved 100\% accuracy with high confidence, citing historical base rates. 
For the Post-cutoff event (Israel-Lebanon), DeepSeek-R1 applied the same Fermi-style decomposition yet produced a miscalibrated prediction. 
The reasoning \textit{appears} equally rigorous in both cases---the 
difference is that Pre-cutoff ``reasoning'' likely \textit{rationalizes} a known outcome rather than deriving it.

\paragraph{Case Study 2: International Agreements.}
Gemini-2.5-Flash achieved a perfect Brier score on the Pre-cutoff NATO question by recalling specific ratification dates---information that should be unavailable under the cutoff instruction. On the structurally identical Post-cutoff question (Artemis Accords), it reverted to statistical heuristics (Poisson distribution). The heuristic is logically sound but reflects genuine uncertainty; the Pre-cutoff precision reflects leaked knowledge.

\paragraph{Key Observation.}
The critical pattern is not that Post-cutoff reasoning fails---uncertainty is expected for genuinely unknown events. Rather, Pre-cutoff reasoning 
\textit{succeeds too well}: models produce confident, accurate predictions with clean traces, suggesting the reasoning is post-hoc rationalization of implicitly accessed outcomes. This is the ``illusion of reasoning''---surface-level logic that masks underlying knowledge retrieval.

\section{Discussion}
\label{sec:discussion}

\paragraph{Summary.}
Across all interventions, SI fails to approximate TI: cutoff instructions close only 48\% of the gap, CoT narrows but does not eliminate it, and reasoning-optimized models exhibit the largest gaps despite the cleanest traces.
Our robustness checks confirm this reflects knowledge leakage, not question difficulty.
The common failure mode is that all interventions target what models \textit{say}, not what they \textit{know}.

\paragraph{Domain patterns.}
The SI--TI gap varies substantially across domains: geopolitical events show the largest gaps (0.111 Brier difference) while business metrics show the smallest (0.010).
We hypothesize this reflects \textit{memorability asymmetry}: high-coverage events like elections become deeply encoded and harder to suppress than lower-salience outcomes.

\paragraph{Reasoning-optimized models.}
The finding that reasoning-optimized models exhibit larger SI--TI gaps despite cleaner traces is predictable given RL training dynamics.
RL optimizes for both instruction-following traces and correct answers; when the model has encoded an outcome, these objectives conflict.
Training resolves this by teaching models to construct plausible reasoning that avoids explicit violations while guiding toward memorized answers---appearing compliant without being compliant.

\paragraph{Implications for reasoning faithfulness.}
Our findings contribute to growing concerns about CoT faithfulness \citep{turpin2023unfaithful,lanham2023measuringfaithfulness}.
We provide forecasting-specific evidence: when models have memorized outcomes, reasoning becomes post-hoc rationalization rather than genuine derivation.
\citep{chen2025reasoningmodelsdontsay} show that reasoning models frequently use hints without verbalizing reliance; our results are consistent---models construct convincing justifications for predetermined answers.
Unlike mathematical tasks where errors can be verified, forecasting offers no ground-truth reasoning path to expose this.

\paragraph{Implications for evaluation.}
SI-based retrospective forecasting represents a particularly insidious form of benchmark contamination: models have seen \textit{outcomes} that inform predictions even when questions are novel \citep{sainz-etal-2023-nlp,balloccu-etal-2024-leak}.
This motivates dynamic benchmarks like ForecastBench \citep{karger2025forecastbench} that restrict evaluation to genuinely future events.
Our results provide empirical justification: prompt-based temporal constraints cannot substitute for genuine temporal separation.

\paragraph{Limitations.}
We cannot definitively rule out residual difficulty confounds, though robustness checks argue against this.
Our trace audits may miss subtle leakage through word choice or framing.
We test only prompt-based interventions; whether parameter-level approaches (e.g., machine unlearning) could achieve better SI fidelity remains open \citep{liu2024rethinking}.
Finally, our evaluation focuses on binary questions; other formats may differ.

\paragraph{Recommendations.} 
\textbf{1. Avoid pre-cutoff evaluation.} Restrict to questions genuinely postdating all models' cutoffs, or use continuously refreshed benchmarks. 
\textbf{2. Do not trust clean traces.}
Performance-based metrics (SI--TI gaps) are more reliable than trace audits for detecting leakage.

\section{Conclusion}
\label{sec:conclusion}

We test whether Simulated Ignorance (SI)---prompting models to ignore post-cutoff knowledge---can substitute for True Ignorance (TI) in retrospective forecasting evaluation. 
Across cutoff instructions, chain-of-thought prompting, and reasoning-optimized models, SI consistently fails: models retain a systematic advantage on pre-cutoff questions even when their reasoning traces appear temporally compliant. Robustness checks confirm this gap reflects knowledge leakage, not question difficulty.
We conclude that SI-based retrospective evaluation is flawed and recommend restricting LLM forecasting benchmarks to post-cutoff events.

\clearpage

\bibliographystyle{named}
\bibliography{ijcai26}

\end{document}